\documentclass[sigconf]{aamas}

\usepackage{balance} 
\usepackage{algorithm}
\usepackage{algpseudocode}
\usepackage{tabularx}
\usepackage{makecell,booktabs,multirow}
\usepackage{listings}
\usepackage{appendix}

\acmSubmissionID{<<182>>}

\title{Subgoal Graph-Augmented Planning for LLM-Guided Open-World Reinforcement Learning}

\author{Shanwei Fan}
\affiliation{
  \institution{The Key Laboratory of Cognition and Decision Intelligence for Complex Systems, Institute of Automation, Chinese Academy of Sciences}
  \city{Beijing}
  \country{China}
  }
  \affiliation{
  \institution{School of Artificial Intelligence, University of Chinese Academy of Sciences}
  \city{Beijing}
  \country{China}
  }
\email{fanshanwei2024@ia.ac.cn}

\author{Bin Zhang}
\affiliation{
  \institution{The Key Laboratory of Cognition and Decision Intelligence for Complex Systems, Institute of Automation, Chinese Academy of Sciences}
  \city{Beijing}
  \country{China}
  }
  \affiliation{
  \institution{School of Artificial Intelligence, University of Chinese Academy of Sciences}
  \city{Beijing}
  \country{China}
  }
\email{zhangbin2020@ia.ac.cn}

\author{Zhiwei Xu}
  \affiliation{
  \institution{School of Artificial Intelligence, Shandong University}
  \city{Jinan, Shandong}
  \country{China}
  }
\email{zhiwei_xu@sdu.edu.cn}

\author{Yingxuan Teng}
\affiliation{
  \institution{The Key Laboratory of Cognition and Decision Intelligence for Complex Systems, Institute of Automation, Chinese Academy of Sciences}
  \city{Beijing}
  \country{China}
  }
  \affiliation{
  \institution{School of Artificial Intelligence, University of Chinese Academy of Sciences}
  \city{Beijing}
  \country{China}
  }
\email{tengyingxuan2024@ia.ac.cn}

\author{Siqi Dai}
\affiliation{
  \institution{The Key Laboratory of Cognition and Decision Intelligence for Complex Systems, Institute of Automation, Chinese Academy of Sciences}
  \city{Beijing}
  \country{China}
  }
  \affiliation{
  \institution{School of Artificial Intelligence, University of Chinese Academy of Sciences}
  \city{Beijing}
  \country{China}
  }
\email{daisiqi2025@ia.ac.cn}

\author{Lin Cheng}
\affiliation{
  \institution{The Key Laboratory of Cognition and Decision Intelligence for Complex Systems, Institute of Automation, Chinese Academy of Sciences}
  \city{Beijing}
  \country{China}
  }
  \affiliation{
  \institution{School of Artificial Intelligence, University of Chinese Academy of Sciences}
  \city{Beijing}
  \country{China}
  }
\email{chenglin2025@ia.ac.cn}

\author{Guoliang Fan}
\affiliation{
  \institution{The Key Laboratory of Cognition and Decision Intelligence for Complex Systems, Institute of Automation, Chinese Academy of Sciences}
  \city{Beijing}
  \country{China}
  }
\email{guoliang.fan@ia.ac.cn}

\begin{abstract}
Large language models (LLMs) offer strong high-level planning capabilities for reinforcement learning (RL) by decomposing tasks into subgoals. However, their practical utility is limited by poor planning-execution alignment, which reflects a critical gap between abstract plans and actionable, environment-compatible behaviors. This misalignment arises from two interrelated limitations: (1) LLMs often produce subgoals that are semantically plausible but infeasible or irrelevant in the target environment due to insufficient grounding in environment-specific knowledge, and (2) single-LLM planning conflates generation with self-verification, resulting in overconfident yet unreliable subgoals that frequently fail during execution. To address these challenges, we propose Subgoal Graph-Augmented Actor-Critic-Refiner (SGA-ACR), a framework that integrates an environment-specific subgoal graph and structured entity knowledge with a multi-LLM planning pipeline that explicitly separates generation, critique, and refinement to produce executable and verifiable subgoals. A subgoal tracker further monitors execution progress, provides auxiliary rewards, and adaptively updates the subgoal graph to maintain alignment between plans and actions. Experimental results on 22 diverse tasks in the open-world game "Crafter" demonstrate the effectiveness of our proposed method.

\end{abstract}


\newcommand{\BibTeX}{\rm B\kern-.05em{\sc i\kern-.025em b}\kern-.08em\TeX}

\begin{document}


\pagestyle{fancy}
\fancyhead{}


\maketitle 


\section{Introduction}

Large Language Models (LLMs) have demonstrated impressive capabilities across a broad spectrum of natural language processing tasks. The successful application of LLMs in these tasks has generated significant interest in leveraging LLMs as planning modules that facilitate task decomposition~\cite{huang2024understanding}. Previous studies have shown that LLMs can decompose complex objectives into high-level plans composed of multiple subgoals, achieving promising results~\cite{wei2022chain,yao2023react,wang2023plan}. Building on these advancements, recent studies have investigated the integration of LLM-based planning into reinforcement learning (RL)~\cite{zhang2024adarefiner,zhou2024large,chen2025causal}. Classic RL methods often suffer from low sample efficiency, particularly in sparse-reward and multitask settings such as open-world environments. In contrast, LLMs, with extensive knowledge and strong planning capabilities, can provide powerful prior knowledge to guide the exploration and decision making of RL agents through high-level plan generation. 

However, applying LLM-based planning to guide RL faces two fundamental challenges that significantly limit its effectiveness. The first challenge is the misalignment of environmental knowledge. LLMs may not possess accurate environment-specific knowledge such as environment dynamics and constraints~\cite{tantrue}. This misalignment may lead LLMs to propose subgoals that violate environmental constraints. The second challenge is the reliability gap created by single-LLM planning, where the same model generates, self-evaluates, and self-repairs its plans. This paradigm amplifies shared biases and is ineffective in revealing the model's own errors, resulting in overconfident but flawed subgoal sequences, which undermines the quality and reliability of planning~\cite{valmeekam2023can, stechlyself}.

Previous works have explored various methods to address these challenges. Some studies fine-tune LLM planners through supervised fine-tuning (SFT) or reinforcement learning from human feedback (RLHF)~\cite{erdoganplan, mohammadi2025learning}. However, SFT-based methods typically require environment-specific planning datasets, which are often unavailable and prohibitively costly to construct manually, while RLHF-based methods are costly and difficult to train, especially in the open-world environment. Other works incorporate Retrieval-Augmented Generation (RAG) into planning~\cite{fan2022minedojo, yoo2024exploratory}, but struggle to combine external knowledge with complex reasoning. Another line of work introduces multi-LLM paradigms~\cite{zhang2024controlling,dongenhancing}, yet these methods suffer from high inference costs and lack dynamic adaptation through environmental feedback. In general, existing methods either rely on parameter fine-tuning or fail to achieve effective alignment among the environment, planning, and execution.

In this paper, our goal is to improve the planning capabilities
of LLMs by simultaneously addressing the two challenges above without the need for parameter fine-tuning, enabling the generation of high-quality plans to assist RL agents. To this end, we propose Subgoal Graph-Augmented Actor-Critic-Refiner (SGA-ACR), a novel framework that seamlessly integrates an environment-specific subgoal graph and a structured entity knowledge base with an LLM-based actor-critic-refiner planning pipeline to generate plans that effectively guide the RL agent. Our framework consists of two stages: an offline stage and an online stage. In the offline stage, structured environmental knowledge is derived from background information using LLMs. In the online stage, the multi-LLM planning module leverages structured knowledge retrieval to access environment-specific knowledge and employs the actor-critic-refiner paradigm for online planning, generating high-level plans that are subsequently embedded into the RL policy. To further align planning with execution, we design a subgoal tracker that provides extra rewards for subgoals achieved in the current plan and adaptively updates the weights of the subgoal graph. Experimental results on 22 diverse tasks in the open-world game "Crafter"~\cite{hafnerbenchmarking} demonstrate the efficacy of our proposed method in improving the planning capabilities of LLMs to effectively guide the exploration and decision-making of the RL agent.

Our key contributions are summarized as follows: (1) We construct the subgoal graph and entity knowledge base from environmental background information, enabling LLMs to access environmental knowledge and perform structured reasoning. (2) We introduce a multi-LLM planning pipeline, which significantly improves planning quality without the need for fine-tuning LLM parameters and remains applicable to any standard LLMs. (3) A subgoal tracker is designed to establish bidirectional feedback between the RL agent and LLM-based planning, further aligning planning with execution.

\section{Related Work}
\subsection{LLMs for Planning}
Recent advances in LLMs have enabled the rise of autonomous planning systems, which exhibit promising planning capacity~\cite{ICLR2024_d53538ba,wu2024can,zhu2025knowagent}. A key application of LLM-based planning is task decomposition, where complex tasks are broken down into manageable subgoals. Existing methods for task decomposition generally follow two paradigms: decomposition-first, where a complete plan is generated before execution~\cite{shen2023hugginggpt,wang2023plan}, and interleaved decomposition, where short-horizon plans are incrementally generated based on the current state~\cite{wei2022chain,yao2023react}. Our framework adopts the latter, dynamically updating plans during interaction, and further integrates LLM-based planning with RL to improve decision making.

\subsection{LLMs for RL}
The advances of LLMs have led to a surge of research that integrates them into RL~\cite{schoepp2025evolving}. Some studies employ LLMs as agents and fine-tune them using RLHF methods~\cite{carta2023grounding,tantrue}. Other studies instead employ LLMs as planners, generating high-level plans to guide RL agents. Among them, AdaRefiner~\cite{zhang2024adarefiner} uses LLMs to generate subgoals and refine them via feedback, but does not address the misalignment between environment and LLMs. Causal-aware LLMs~\cite{chen2025causal} incorporate learned causal graphs, but require curated validation environments. In contrast, our method extracts a complete subgoal graph offline from background information, eliminating the need for online learning and validation, and employs a multi-LLM planning process to improve robustness.

\subsection{RAG-based LLMs}
RAG techniques enhance LLM generation by retrieving task information from a vectorized document database as context~\cite{lewis2020retrieval}. Graph-based RAG techniques further organize knowledge into graphs, enabling reasoning over relational structures~\cite{han2025retrieval}. Recent studies have applied these techniques in LLM-based game agents: AVA~\cite{ma2025ava} retrieves unit information for tactical guidance, and GoGs~\cite{leung2025knowledge} builds goal-oriented graphs for subgoal retrieval. Similarly, we construct an environment-specific subgoal graph and an entity knowledge base. In contrast, rather than merely feeding retrieved knowledge into a single LLM, we integrate structured knowledge retrieval into a multi-LLM planning paradigm, tailoring the retrieved knowledge to each role-specific LLM.


\section{PRELIMINARIES}
\subsection{Partially Observable Markov Decision Process}
In most open-world decision-making environments, an agent can only access observations within its field of view rather than the full global state. Therefore, the environment can be modeled as a partially observable Markov decision process (POMDP)~\cite{sondik1971optimal}. The POMDP can be defined as a tuple $(\mathcal{S}, \mathcal{A}, \mathcal{P}, \Omega, \mathcal{O}, R, \gamma).$
Here, $s \in \mathcal{S}$ represents the state of the environment and $a \in \mathcal{A}$ denotes an action sampled from a policy $\pi$. The state transition probability $\mathcal{P}(s' \mid s, a)$ represents the environment dynamics, where $s'$ is the next state following action $a$ from state $s$. The observation $o \in \Omega$ is obtained through the observation function
$\mathcal{O}(o \mid s, a),$ which provides the probability of $o$ given the next state $s'$ and the action $a$. The reward function $R(s, a)$ quantifies the reward attained by executing action $a$ in state $s$. Finally, $\gamma$ is the discount factor, which modulates the significance of future rewards in present decision-making. The goal of the agent is to maximize the discount
return $R_t = \sum_{k=0}^{\infty}\gamma^{k} r_{t+k}.$

\begin{figure*}[t]
    \centering
    \includegraphics[width=\textwidth]{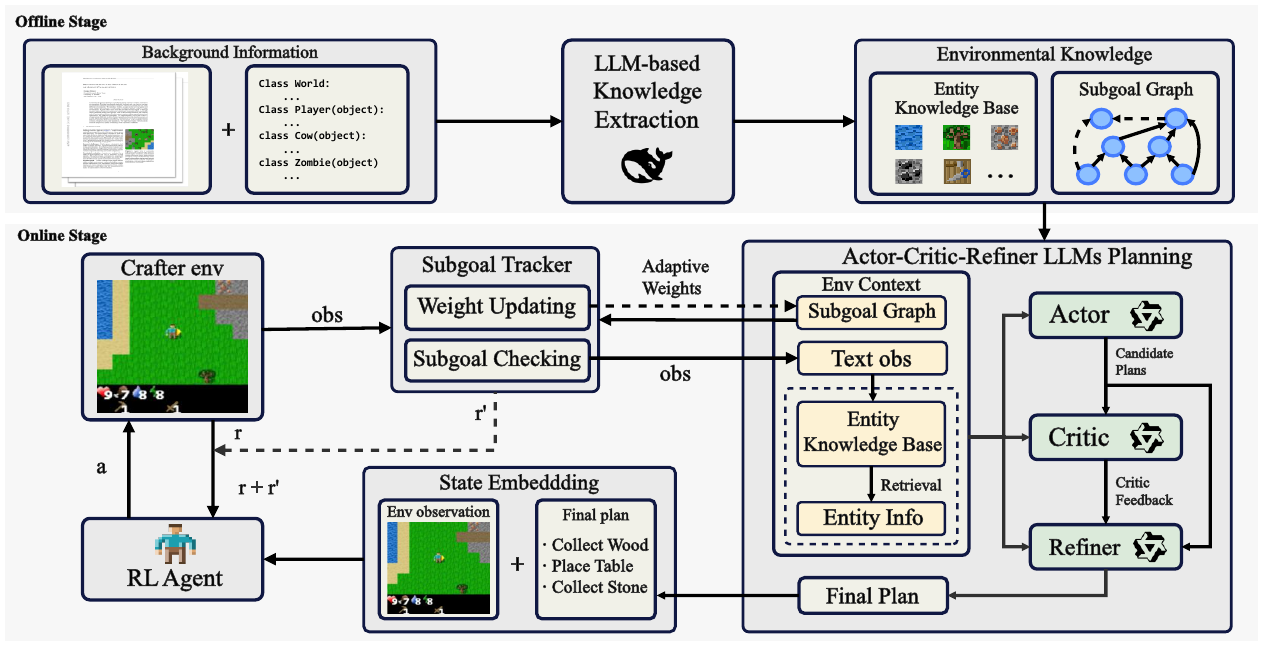}
    \caption{Framework of our SGA-ACR. In the offline stage, the LLM extracts structured knowledge from background information(in the subgoal graph, dashed lines and solid lines denote OR-edges and AND-edges, respectively). In the online stage, the RL agent optimizes its policy through interaction with the environment, and the multi-LLM planning module generates plans to guide exploration and decision-making. The subgoal tracker coordinates the planning module and the RL agent.}
    \label{fig:full_framework}
    \Description{framework fig}
\end{figure*}

\subsection{Goal-Conditioned RL}
In our framework, the LLMs generate the high-level plan consisting of multiple subgoals, each representing a discrete intermediate objective in the task hierarchy, which serves as guidance during the RL agent's execution phase. Based on this definition, we augment the POMDP with a goal space $\mathcal{G}$, where each goal $g \in \mathcal{G}$ corresponds to a plan in natural language generated by the LLM. Thus, the decision-making algorithm can be formulated as goal-conditioned RL~\cite{liu2022goal}. In this formulation, the goal-conditioned policy $\pi(a \mid o, g)$ maps observations and a goal to an action. $R(s, a, g)$ is the goal-conditioned reward function. The goal-conditioned value functions are:
\begin{equation}
V^{\pi}(s,g) = \mathbb{E}_{\pi}\!\left[\sum_{k=0}^{\infty}\gamma^{k} r_{t+k} \,\bigg|\, s_t=s, g\right],
\end{equation}

\begin{equation}
    Q^{\pi}(s,a,g) = \mathbb{E}_{\pi}\!\left[\sum_{k=0}^{\infty}\gamma^{k} r_{t+k} \,\bigg|\, s_t=s, a_t=a, g\right],
\end{equation}
and the corresponding advantage function is defined as:
\begin{equation}
    A^{\pi}(s, a, g) = Q^{\pi}(s, a, g) - V^{\pi}(s, g).
\end{equation}
Based on the above definitions, we adopt the PPO algorithm~\cite{schulman2017proximal} to train the policy.


\section{Method}
In this section, we present the overall SGA-ACR framework. First, the offline knowledge extraction and construction process is introduced, where a subgoal graph and a structured entity knowledge base are derived from environmental background information. Based on this foundation, the LLM-based actor-critic-refiner planning module operates in the online stage, supported by RAG-based techniques, to generate high-level plans. Next, we describe the subgoal tracker, which bridges the RL agent and the planning module by enabling bidirectional feedback through monitoring subgoal execution progress. Finally, we summarize the general training procedure. The SGA-ACR framework is illustrated in Figure \ref{fig:full_framework}.

\subsection{Knowledge Extraction and Construction}\label{sec:bgkn}
To bridge the knowledge gap between LLMs and the environment, a \textbf{dependency-aware subgoal graph} and a structured \textbf{entity knowledge base} are extracted from environment-specific background information. First, following the method of Spring~\cite{wu2023spring}, we extract textual descriptions from the environment paper as background information. To further capture environment dynamics, we additionally incorporate a subset of the core implementation code of "Crafter"~\cite{hafnerbenchmarking} as supplementary knowledge. Based on this background information, unlike standard RAG methods that construct unstructured text chunks for retrieval or typical graph-based RAG methods that build entity-level knowledge graphs, we construct a dependency-aware graph at the subgoal level together with a structured entity knowledge base at the entity level. The basic idea is that the subgoal graph provides explicit reasoning paths to regulate the planning process, while the entity knowledge base offers precise entity-level information to strengthen the ability of LLMs to interpret observations.

\textbf{Subgoal-level dependency graph.} First, we formalize the subgoal graph as a directed acyclic graph $G(V,E),$ where $V$ denotes the set of nodes and $E \subseteq V \times V$ represents the set of edges from source to target nodes. Each node corresponds to a subgoal and is associated with attributes that include its name, description, preconditions and postconditions. The preconditions specify the prerequisites that are needed before the goal can be pursued, and the postconditions describe the state changes after completing the subgoal. Each edge represents the dependency relation between subgoals. An edge from subgoal $v_i$ to subgoal $v_j$ indicates that $v_i$ is a prerequisite for $v_j$. To capture different dependency relations among preconditions, we define two types of edges: AND-edges, where all prerequisites must be satisfied (e.g., \textit{Collect Wood} and \textit{Place Table} are both required to achieve \textit{Make Wood Sword}); and OR-edges, where any single prerequisite is sufficient. (e.g., either \textit{Make Wood Sword} or \textit{Make Stone Sword} enables \textit{Defeat Zombie}). To construct the subgoal graph, LLMs are leveraged to extract the complete set of subgoals $V_{\text{sub}}$ sufficient to complete the task, together with their associated information $I_{\text{sub}}$:
\begin{equation}
    V_{\text{sub}} \, , \, I_{\text{sub}} \,=\, LLM_{\text{extract}}(Env_{\text{paper}}\, ,\, Env_{\text{code}}),
\end{equation}
where $Env_{\text{paper}}$ and $Env_{\text{code}}$ denote the environment paper and the implementation code in the background information, respectively. Based on the attribute of preconditions in $I_{\text{sub}}$, we derive the edges and construct the subgoal graph. 

\textbf{Entity-level knowledge base.} Information about interactive entities in the environment is extracted and organized into a structured knowledge base $K$, indexed by the name of the entities:
\begin{equation}
    K \,=\, LLM_{\text{extract}}(Env_{\text{paper}}\,,\, Env_{\text{code}}\,, \, V_{\text{sub}} \,, \, I_{\text{sub}}).
\end{equation}
Each entity is associated with attributes including its name, type, description, and related subgoals. During entity knowledge extraction, the subgoal graph information is incorporated as contextual input, enabling the knowledge base to capture the connections between entities and subgoals, thereby facilitating subgoal planning.

In summary, a subgoal graph and an entity knowledge base are constructed from environmental background information in the offline stage, providing structured knowledge that can be retrieved to support LLM-based planning.

\subsection{LLM-based Actor-Critic-Refiner Planning}\label{sec:acr}
To improve the quality of the LLM-based planning, we design a multi-LLM planning framework based on the actor-critic-refiner paradigm. In this framework, structured knowledge retrieval is integrated into the planning process through RAG and graph-based RAG techniques. The core idea of this framework consists of two aspects. First, through a multi-LLM collaborative paradigm rather than single-LLM planning, deficiencies in the initial plans can be effectively identified and corrected, thereby improving the overall quality and reliability of planning. Second, by assigning distinct roles to different LLMs and equipping them with role-specific retrieved knowledge as context, our framework decomposes complex planning into specialized subtasks, allowing each LLM to focus on its specific task while reducing the overall burden. During planning, the information flows sequentially from the actor to the critic and finally to the refiner, ultimately producing the final plan.

\textbf{Actor LLM.} As the first module in the planning process, the actor is responsible for generating the initial candidate plans. In the actor module, entity names are first extracted from the agent's textualized observations. For each observed entity, the corresponding structured knowledge is then retrieved from the entity knowledge base $K$ according to the entity name. Given the entity set $\mathcal{E}$, we formulate the entity retrieval process as:
\begin{equation}
    K(e) = \{n_e\, , \, t_e\, , \,d_e\, , \,s_e\} \quad \forall e \in \mathcal{E},
\end{equation}
where for each entity $e$, $n_e$, $t_e$, and $d_e$ denote its name, type, and description, respectively, while $s_e \subseteq V_{\text{sub}}$ represents the subgoals associated with the entity. To provide the LLM with explicit planning paths, we perform graph-to-text verbalization on the subgoal graph to obtain the verbalized graph $G_{\text{ver}}$. Specifically, subgoals without preconditions are first identified and represented as root nodes. The subgoals dependent on root nodes are then identified and their dependency relations are expressed in textual form. This process is applied iteratively until the subgoal graph is fully verbalized. Subsequently, the verbalized subgoal graph, textualized observations, and entity knowledge are provided to the actor, which generates $k$ candidate plans together with their corresponding rationales. By generating multiple candidate plans, the actor aims to cover potentially optimal plans, while the verbalized subgoal graph guides the actor to focus on path planning over the subgoal graph. The generation process can be formulated as:
\begin{equation}
    \{(p_i,\, q_i)\}_{i=1}^k \,=\, LLM_{\text{actor}}(o_{\text{text}},\;G_{\text{ver}},\;\{K(e)\}),
\end{equation}
where $p_i$ denotes the $i$-th candidate plan, $q_i$ denotes the rationale of the $i$-th candidate plan, and $o_{\text{text}}$ denotes the textualized observations.

\textbf{Critic LLM.} The task of the critic LLM is to evaluate all candidate plans and provide suggestions for refinement. As a key procedure in the planning process, the accuracy of the evaluation directly affects the improvement of the plan. To ensure accurate evaluation, we first adopt the entity linking technique in graph-based RAG to retrieve the structured knowledge of all subgoals contained in the candidate plans, which provides the critic with detailed knowledge required for evaluation. It should be noted that, beyond the subgoals contained in the candidate plans, information about their forward and backward $k$-hop neighboring nodes in the graph may also be retrieved. In "Crafter", we observe that retrieving the structured knowledge of the subgoals in the candidate plans is sufficient for the critic. The retrieval process can be formulated as:
\begin{equation}
    G(v) = \{n_v\, , \, d_v\, , \,pre_v\, , \,post_v\} \quad \forall v \in V,
\end{equation}
where for each subgoal node $v$, $n_v$, $d_v$, $pre_v$ and $post_v$ denote its name, description, preconditions and postconditions, respectively. Next, the critic takes as input the textualized observations, the retrieved entity knowledge, the verbalized subgoal graph, the detailed subgoal information, and the actor's output. Based on these inputs, the critic evaluates each candidate plan and provides corresponding suggestions for refinement. To explicitly differentiate the quality of candidate plans, we further prompt the critic to rank them. In addition, a refinement flag is produced to determine whether the top-ranked plan requires further modification. The overall generation process can be formulated as:
\begin{equation}
\begin{aligned}
    \{f_i\}_{i=1}^k,\; rank,\; flag 
    &= LLM_{\text{critic}}\big( o_{\text{text}},\; G_{\text{ver}},\; \{K(e)\}, \\
    &\quad \{G(v)\},\; \{(p_i, q_i)\}_{i=1}^k \big),
\end{aligned}
\end{equation}
where $f_i$ denotes the feedback of the critic for the $i$-th candidate plan, $rank$ denotes the ranking of the candidate plans and $flag$ denotes the refinement flag.

\textbf{Refiner LLM.} The task of the refiner LLM is to integrate the feedback on the candidate plans and then generate the final plan. Before the refiner operates, we first determine whether refinement is necessary: if the flag is "No", the top-ranked plan is directly adopted as the final plan; if the flag is "Yes", the final plan is generated by the refiner. Note that refinement is not limited to the top-ranked plan; instead, all candidate plans, along with their corresponding feedback, are provided to the refiner. The basic idea is that, beyond correcting deficiencies in the top-ranked plan, the refiner can also integrate the strengths of other candidate plans to produce a better plan. The generation of the final plan is formulated as:
\begin{equation}
    p_{\text{final}} = LLM_{\text{refiner}}(o_{\text{text}},\;G_{\text{ver}},\;\{K(e)\},\;\{p_i,\,f_i\}_{i=1}^k,\;rank),
\end{equation}
where $p_{\text{final}}$ denotes the final plan.

In summary, the actor, critic, and refiner modules work collaboratively to achieve efficient integration of environmental knowledge and planning optimization, ensuring the generation of reliable and high-quality plans.

\subsection{Subgoal Tracker}\label{sec:tra}
\begin{figure}[t]
    \centering
    \includegraphics[width=1.0\linewidth]{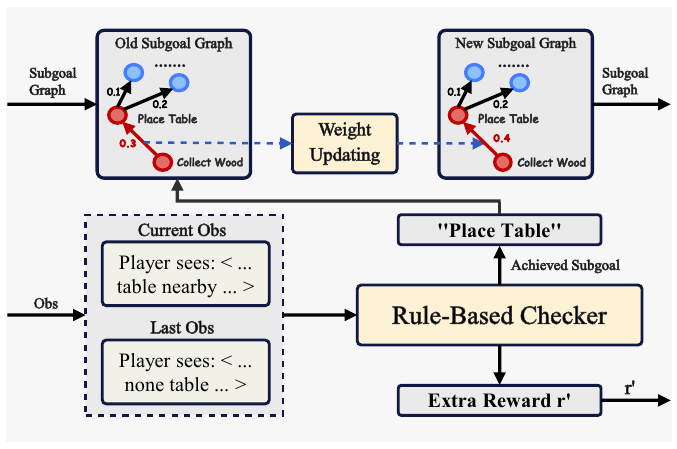}
    \caption{An example of Subgoal Tracker. Based on the state change, the checker identifies that the agent has achieved \textit{Place Table}. The corresponding edge is then updated with a new weight, while an extra reward is assigned to the agent.}
    \label{fig:tracker}
    \Description{tracker fig}
\end{figure}

While the plans generated by the planning module can be integrated into the RL agent for guidance, a misalignment between planning and execution still remains. The misalignment arises from two aspects. First, during the decision-making process, the RL agent cannot receive feedback on whether its execution follows the generated plan. Second, during planning, the planning module has no access to the agent's learning progress and its ability to interpret the plan. To address the misalignment, we design a subgoal tracker, a module that establishes bidirectional feedback between the planning module and the RL agent, enabling better coordination between them. This module consists of three components: subgoal completion detection, subgoal completion reward, and graph weight update. Figure \ref{fig:tracker} illustrates an example of the subgoal tracker's workflow.

\textbf{Subgoal completion detection.} In Section \ref{sec:bgkn}, the postconditions for each subgoal were extracted, which reflect the state changes of the agent after the subgoal is achieved. Based on these postconditions, we design a rule-based subgoal checker. Specifically, at environment step $t$, the checker extracts the set of all objects and their corresponding values from the current textualized observations $o^{t}_{\text{text}}$ and the previous observation $o^{t-1}_{\text{text}}$. Let
\begin{equation}
    \mathcal{O}^{t} = \{ (x, v^{t}(x)) \mid x \in \text{Obs}(o^{t}_{\text{text}}) \}
\end{equation}
denote the set of objects $x$ and their values $v^{t}(x)$ observed at step $t$, where $\text{Obs}(\cdot)$ is the object extraction function. Objects present in both steps are compared based on their value differences, while objects appearing in only one step are marked as either \textit{appear} or \textit{disappear}. Thus, the state change at step $t$ is represented as:
\begin{equation}
\begin{aligned}
    \Delta^{t}
    \,=\,& \{\, (x,\, v^{t}(x) - v^{t-1}(x)) \mid x \in \mathcal{O}^{t} \cap \mathcal{O}^{t-1} \,\} \\
    &\;\cup\; \{\, (x,\, \text{appear}) \mid x \in \mathcal{O}^{t} \setminus \mathcal{O}^{t-1} \,\} \\
    &\;\cup\; \{\, (x,\, \text{disappear}) \mid x \in \mathcal{O}^{t-1} \setminus \mathcal{O}^{t} \,\}.
\end{aligned}
\end{equation}
The detected state change $\Delta^{t}$ is then compared with the postconditions of each subgoal in the current plan; if they match, the subgoal is considered achieved.

\begin{algorithm}[t]
\caption{Pseudo Code for SGA-ACR}\label{alg:training}
\begin{algorithmic}[1]
\Require policy $\pi_\theta(a \mid o,\,p_{\text{emb}})$; text embedding function $f_{emb}(\cdot)$; LLM-based planning module $\textproc{PlanGenerate}$; subgoal tracker $\textproc{Tracker}$; subgoal graph $G$; entity knowledge base $K$; planning interval $H$; max training steps $N$; batch size $M$
\Ensure Policy $\pi_\theta$
\State Initialize policy params $\theta$, replay buffer $\mathcal{B}$, subgoal-tracker state; graph weights; set $o_0 \leftarrow \text{env.reset()}$
\For{$t=0$ \textbf{to} $N$}
    \If{$t\;\% \;H=0 \;\; \mathbf{or}\;\; \forall sg_i \in p \;\text{is achieved}$} \Comment{LLM-based planning process}
      \State $p \leftarrow \textproc{PlanGenerate}(o_t,\,G,\, K)$
      \State \textproc{Tracker.NewPlan}$(p)$ \Comment{update the planned counts}
    \EndIf
    \State Sample $a_t \sim \pi_\theta(\cdot \mid o_t,\, f_{emb}(p))$
    \State $o_{t+1}\leftarrow \text{env.step($a_t$)}$
    \State $r'_t , \;G\leftarrow \textproc{Tracker.Step}(p,\, o_t,\, o_{t+1},\, G)$ \Comment{check the subgoals and update the graph weights}
    \State Push $(o_t,\,p,\,a_t,\,r_t\,{+}\,r'_t,\,o_{t+1})$ into $\mathcal{B}$
    \If{$(t+1)\;\% \;M=0$}
      \State $\theta \leftarrow \textproc{PPO.Update}(\mathcal{B})$; \textproc{Clear}$(\mathcal{B})$
    \EndIf
    \State $o_t \leftarrow o_{t+1}$
\EndFor
\end{algorithmic}
\end{algorithm}

\textbf{Subgoal completion reward.} An extra reward is provided to the RL agent when a subgoal is achieved. However, since a subgoal may be achieved multiple times within a plan, we avoid assigning extra rewards to the same subgoal more than once during a single plan, as it could lead the agent to execute that subgoal repeatedly and eventually diminish the exploration of more complex behaviors. Hence, we provide the extra reward only when each subgoal in the plan is achieved for the first time. The calculation of the extra reward at step $t$ is defined as:
\begin{equation}
    r'_t = \sum_{i=1}^n \alpha \cdot 
    \begin{cases}
        1 & \text{if } sg_i \text{ is achieved for the first time} \\
        0 & \text{otherwise}
    \end{cases}
\end{equation}
where $\alpha$ is the hyperparameter that controls the scale of the extra rewards, $sg_i$ represents the $i$-th subgoal in the plan, and $n$ represents the total number of subgoals in the plan. Through extra rewards, the RL agent receives explicit incentives to follow the plan during decision making.

\textbf{Graph weight update.} To enable the LLMs to receive feedback from the RL agent, the weights in the subgoal graph are adaptively updated. Each weight is defined as the success rate of its corresponding subgoal. Specifically, during training, we maintain two counts for each subgoal $v_i$: its achieved count $N_i^a$ and its planned count $N_i^p$. Thus, the success rate of subgoal $v_i$ is computed as: $\omega(v_i) = N_i^a/N_i^p$. For subgoals without prerequisites, their success rates are represented by the weights of root nodes in the graph, whereas for subgoals with prerequisites, the success rates are represented by the weights of the incoming edges. Moreover, for each subgoal, the weights of the AND-edges are shared since all AND-edges must be satisfied jointly to achieve the subgoal. In contrast, the weights of the OR-edges are independent, as satisfying any single edge is sufficient to achieve the subgoal. Based on the definition, the update of graph weights depends on two aspects. On the one hand, whenever a new plan is generated, we update the planned count of each subgoal included in the plan: $N_i^p \leftarrow N_i^p + 1$. On the other hand, whenever the subgoal checker detects that a subgoal is achieved, we update the achieved count for the subgoal: $N_i^a \leftarrow N_i^a + 1$. During planning, the LLMs in the planning module access these success rates to prioritize subgoals with higher success rates, effectively creating a curriculum that gradually progresses from easier to more challenging subgoals.

\subsection{Training Procedure}
The training process of our framework follows the standard RL paradigm. Specifically, at each training step, the RL agent receives a suggested plan $p$, consisting of three subgoals, from the planning module. This plan is then incorporated into the agent's policy $\pi(a \mid o, \,p_{emb})$ for the interaction with the environment, where $p_{emb}$ denotes the text embedding of plan $p$. In our implementation, we employ SentenceBERT~\cite{reimers2019sentence} to compute the text embeddings. To balance computational cost with the nature of open-world environments, the planning module is queried at fixed intervals or after all subgoals in the current plan are achieved, rather than at every step. In addition, the subgoal tracker operates throughout the training process to ensure coordination between planning and execution. The training procedure is summarized in Algorithm \ref{alg:training}.


\section{Experiment}
In this section, we conduct experiments in the \textit{Crafter} environment~\cite{hafnerbenchmarking} to evaluate the effectiveness of SGA-ACR by answering the following questions: (1) Can SGA-ACR outperform other decision-making methods in the Crafter environment? (2) Does SGA-ACR demonstrate robust performance across LLMs of different parameter scales? (3) How effective is each component of SGA-ACR and what is its contribution to overall performance?

\subsection{Experiment Setup}

\noindent\textbf{Environment description.} Crafter is a 2D version of Minecraft, with a game map size of 64×64 and a player’s field of view of 9×7. The environment contains various interactive objects including various resources (e.g., trees, stones), hostile entities (e.g., zombies), and craftable items. The player’s goal is to unlock as many achievements as possible. The open-world nature of Crafter, where the agent is expected to master a wide range of skills to accomplish 22 diverse tasks, including collecting resources, crafting tools, and surviving against environmental hazards, makes it particularly suitable for evaluating our framework's ability to generate high-quality plans and effectively guide RL agents.

\begin{table}[t]
  \caption{Performance comparison between SGA-ACR and baselines in terms of score and reward metrics (mean $\pm$ std).}
  \label{tab:results}
  \centering
  \begin{tabularx}{\linewidth}{c l c c}
    \toprule
    \makecell{Method\\Type} & Method & Score (\%) & Reward \\
    \midrule
    \multirow{2}{*}{Ours}
      & SGA-ACR(@5M) & \textbf{29.6} $\pm$ 1.1 & \textbf{13.3} $\pm$ 1.2 \\
      & SGA-ACR(@1M) & \textbf{17.6} $\pm$ 1.2 & \textbf{11.1} $\pm$ 1.3 \\
    \midrule
    \multirow{4}{*}{\makecell[c]{LLM-guided\\RL methods}}
      & Causal-aware(@5M) & 27.7 $\pm$ 1.4 & 12.7 $\pm$ 1.3 \\
      & Causal-aware(@1M) & 13.2 $\pm$ 1.5 &  10.3 $\pm$ 1.4 \\
      & AdaRefiner(@5M)   & 26.8 $\pm$ 1.7 & 12.5 $\pm$ 1.5 \\
      & AdaRefiner(@1M)   & 14.3 $\pm$ 1.6 &  10.1 $\pm$ 1.3 \\
    \midrule
    \multirow{3}{*}{\makecell[c]{LLM-based\\methods}}
      & SPRING(w/GPT-4)   & 27.3 $\pm$ 1.2 & 12.3 $\pm$ 0.7 \\
      & Reflexion(w/GPT-4)& 11.7 $\pm$ 1.4 & 10.3 $\pm$ 1.3 \\
      & ReAct(w/GPT-4)    &  8.3 $\pm$ 1.2 &  7.4 $\pm$ 0.9 \\
    \midrule
    \multirow{4}{*}{\makecell[c]{RL\\methods}}
      & PPO(ResNet)(@1M) & 14.9 $\pm$ 1.4 & 10.2 $\pm$ 1.3 \\
      & DreamerV3(@1M)    & 14.5 $\pm$ 1.6 & 11.7 $\pm$ 1.9 \\
      & PPO(@1M)          & 11.3 $\pm$ 1.2 &  9.5 $\pm$ 0.9 \\
      & Rainbow(@1M)      &  4.3 $\pm$ 0.2 &  5.0 $\pm$ 1.3 \\
    \midrule
    \multirow{2}{*}{\makecell[c]{Additional\\refs}}
      & Human Experts & 50.5 $\pm$ 6.8 & 14.3 $\pm$ 2.3 \\
      & Random        &  1.6 $\pm$ 0.0 &  2.1 $\pm$ 1.3 \\
    \bottomrule
  \end{tabularx}
\end{table}

\noindent\textbf{Evaluation Metrics.} In Crafter, the performance of an agent is evaluated using three metrics: reward, success rate, and score. The reward is designed to reflect the agent's skills. Each time the agent unlocks a new achievement, it receives a reward of $+1.$ In addition, the agent is rewarded with $+0.1$ or penalized with $-0.1$ for every gain or loss of a health point, respectively. The success rate is defined as the proportion of training episodes in which the agent completes an achievement, reflecting the breadth of abilities acquired. Multiple completions of the same achievement within a single episode do not increase the success rate. The score measures the agent's proficiency across skills by averaging the success rates ($s_i \in [0,100]$) of all 22 achievements in log-space (geometric mean): $\text{Score} = \exp\!\left( \frac{1}{N} \sum_{i=1}^{N} \ln(1+s_i) \right) - 1,$ where $N=22$ is the total number of achievements.

\begin{figure}[t]
    \centering
    \includegraphics[width=1.0\linewidth]{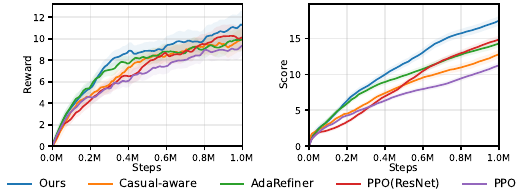}
    \caption{Comparison of reward and score learning curves.}
    \label{fig:learning}
    \Description{learning curve}
\end{figure}

\begin{figure*}[ht]
    \centering
    \includegraphics[width=\textwidth]{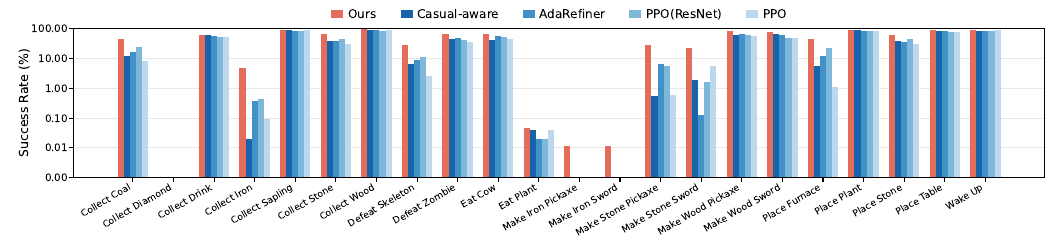}
    \caption{Success rates of unlocking 22 different achievements in log scale (at 1M training steps).}
    \label{fig:scr}
    \Description{scr fig}
\end{figure*}

\begin{figure}[ht]
    \centering
    \includegraphics[width=1.0\linewidth]{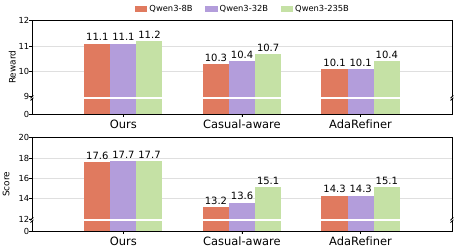}
    \caption{Performance of SGA-ACR compared with LLM-guided RL Baselines across various model scales.}
    \label{fig:scale}
    \Description{scale fig}
\end{figure}

\noindent\textbf{Baselines.} The baselines in our experiments can be grouped into four categories. (1) LLM-guided RL methods: This category includes AdaRefiner~\cite{zhang2024adarefiner} and Causal-aware LLMs~\cite{chen2025causal}, which employ LLMs as planners to provide high-level guidance to RL agents. (2) LLM-based methods: This category covers SPRING~\cite{wu2023spring}, Reflexion~\cite{shinn2023reflexion}, and ReAct~\cite{yao2023react}, where LLMs act as agents to make decisions in the environment. (3) RL methods: This category consists of PPO (ResNet version)~\cite{moon2023discovering}, PPO~\cite{schulman2017proximal}, Rainbow~\cite{hessel2018rainbow}, and DreamerV3~\cite{hafner2023mastering}, which rely purely on RL algorithms to interact with the environment and learn policies. (4) Additional references: This category includes human experts~\cite{hafnerbenchmarking} and a random policy, which serve as upper and lower bounds for additional references.

Qwen3-8B~\cite{yang2025qwen3technicalreport} is adopted as the base LLM in our planning module, and to ensure a fair comparison, the same model is also used in the LLM-guided RL methods. In addition, DeepSeek-V3 (671B)~\cite{liu2024deepseek} is used to extract knowledge in our method. The implementation details and hyperparameters are provided in Appendix \ref{sec:A}.

\subsection{Main Results} \label{sec:mainresult}
In the experimental results, for methods without publicly available implementations, we reference performance metrics from their original publications to ensure consistency in experimental setup and evaluation. 
More detailed results are provided in Appendix \ref{sec:B} and additional case studies are presented in Appendix \ref{sec:C}.

According to Table \ref{tab:results}, our method significantly outperforms all baselines except for the human expert. Specifically, our method outperforms the best RL-based method PPO (ResNet) by 13.5\%, indicating that the high-level plans generated by the LLM-based planning module substantially enhance RL performance in open-world environments. Compared with Reflection and ReAct, our method effectively resolves the issue of misalignment with the environment, leading to significant performance improvements. Although SPRING incorporates environmental information, our method achieves a higher performance of 8.3\%, highlighting the advantage of integrating structured-knowledge-aware planning with RL. Furthermore, our method outperforms AdaRefiner and Causal-aware LLMs by 16.5\% / 20.6\% at 1M steps and 8.4\% / 5.8\% at 5M steps, respectively. Together with Figure \ref{fig:learning}, these results indicate that our method can generate higher-quality plans and achieve better alignment between planning and execution, enabling more efficient learning and superior convergence performance.

Regarding the success rates of different achievements, as illustrated in Figure \ref{fig:scr}, our method achieves the highest success rates in 20 out of 22 tasks, except for two low-level ones, \textit{Wake up} and \textit{Collect Sapling}, which are only 1.2\% and 1.4\% lower than the highest values, respectively. In particular, our method shows substantial improvements in unlocking deeper achievements(e.g., \textit{Collect Iron}), and is the only method capable of unlocking \textit{Make Iron Pickaxe} and \textit{Make Iron Sword} in 1M steps. This indicates that the planning module benefits from the integration of structured knowledge with the multi-LLM pipeline, thereby enhancing the effectiveness of the generated plans. Meanwhile, extra reward feedback enables the RL agent to better interpret and follow the generated plans, thereby progressively unlocking higher-level achievements.

\subsection{Model-Scale Analysis} \label{sec:modelscale}
To investigate the impact of the LLM parameter scale, we conduct experiments using Qwen3-32B and Qwen3-235B as the base LLM for both our method and the LLM-guided RL baselines. As shown in Figure \ref{fig:scale}, our method remains robust across model scales, while both AdaRefiner and Causal-aware LLMs exhibit clear sensitivity, with their performance improving as model capacity increases. Based on these results, we offer the following analysis: (1) In AdaRefiner, the LLM planner needs to accurately interpret reflections on the execution of the previous plan and the agent's comprehension score, while reasoning entirely based on its own knowledge, thereby making the quality of planning highly dependent on model capacity. (2) Causal-aware LLMs rely on the intrinsic causal reasoning ability of the LLM, as they require causal discovery from the agent's observations and the generation of subgoals from entity-level causal graphs; (3) Our method introduces a subgoal graph that regularizes the planning process to feasible paths and decomposes the overall planning task into a sequence of simple subtasks handled by role-specific LLMs. This design obviates the need for LLMs to possess strong intrinsic capabilities, making our method robust to variations in model scale and broadly applicable to existing LLMs.

\begin{table}[t]
  \centering
  \caption{Ablation studies of SGA-ACR}
  \label{tab:ablation}
  \begin{tabular}{lcc}
    \toprule
    Method(@1M) & Score (\%) & Reward \\
    \midrule
    SGA-ACR        & \textbf{17.6} $\pm$ 1.2 & \textbf{11.1} $\pm$ 1.3 \\
    \midrule
    text-RAG         & 15.8 $\pm$ 1.4     & 10.2 $\pm$ 1.4    \\
    \textit{w/o} graph   & 14.8 $\pm$ 1.3  & 9.8 $\pm$ 1.3\\
    \textit{w/o} entity info  & 16.4 $\pm$ 1.0 & 10.6 $\pm$ 1.2 \\
    \textit{w/o} environmental info    & 13.1 $\pm$ 1.8    & 9.7 $\pm$ 1.5    \\
    \midrule
    actor-only  & 15.4 $\pm$ 1.4 & 10.4 $\pm$ 1.5\\
    random plan      & 14.0  $\pm$ 1.4   & 9.9  $\pm$ 1.5\\
    \textit{w/o} refiner & 15.3 $\pm$ 1.5    & 10.5 $\pm$ 1.3   \\
    \textit{w/o} critic  & 13.9 $\pm$ 1.3    & 10.1 $\pm$ 1.6    \\
    \textit{w/o} flag    & 16.7 $\pm$ 1.1    & 10.6 $\pm$ 1.2    \\
    \midrule
    extra $r$ every time & 15.5 $\pm$ 1.4 & 10.0 $\pm$ 1.2\\
    \textit{w/o} extra $r$        & 16.1 $\pm$ 1.1 & 10.5 $\pm$ 1.3\\
    \textit{w/o} weight update & 16.8 $\pm$ 0.8 & 10.8 $\pm$ 1.1\\
    \bottomrule
  \end{tabular}
\end{table}

\subsection{Ablation Studies} \label{sec:aba}
To investigate the contributions of each component of SGA-ACR, we conduct comprehensive ablation studies. Our framework is characterized by three key modules: the knowledge module (Section \ref{sec:bgkn}), the planning module (Section \ref{sec:acr}), and the subgoal tracker (Section \ref{sec:tra}). Therefore, we conduct ablation experiments on each of these modules over 1M training steps. The results, summarized in Table \ref{tab:ablation}, demonstrate that SGA-ACR is better than all other variants.

\begin{figure}[t]
    \centering
    \includegraphics[width=1.0\linewidth]{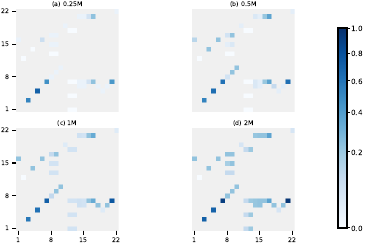}
    \caption{Subgoal graph weights at different training steps in SGA-ACR. Nodes 1-21 correspond to the subgoals from \textit{Collect Coal} to \textit{Place Table} in Figure \ref{fig:scr}, and node 22 represents \textit{Sleep}. Masked entries indicate the absence of relations.}
    \label{fig:hot}
    \Description{hot fig}
\end{figure}

\textbf{Knowledge module ablation.} To examine the contributions of structured knowledge, we design the following experiments: (1) \textit{text-RAG}, in which the subgoal graph and entity knowledge base are removed and replaced by the retrieval of unstructured text chunks containing environmental information based on textualized observations during planning; and (2) \textit{w/o} graph, \textit{w/o} entity info, and \textit{w/o} background info, where the subgoal graph, the entity knowledge base, and both are removed from SGA-ACR, respectively. 

The results of \textit{w/o} background info demonstrate that incorporating environmental knowledge into LLM-based planning is crucial to effectively guide the RL agent. Compared to the \textit{text-RAG} variant, our method achieves significantly better performance, indicating that structured knowledge is more effective than unstructured text chunks, as it explicitly forms reasoning chains that facilitate planning. Furthermore, the performance drop of the \textit{w/o} graph and \textit{w/o} entity info variants indicates that the subgoal graph plays a vital role in improving the planning capability of LLMs, while the entity knowledge base further enhances performance by strengthening the model's understanding of the environment.

\textbf{Planning module ablation.} To investigate the contributions of key components in our planning module, we design the following experiments: (1) \textit{actor-only}, where all environmental knowledge is provided as context to a single LLM that directly generates the final plan; (2) \textit{random plan}, where the actor generates multiple candidate plans and the final plan is selected randomly; (3) \textit{w/o} refiner, where the critic outputs the final plan after evaluation; (4) \textit{w/o} critic, where the refiner directly generates the final plan based on the actor's output, resembling a self-reflection process, and (5)\textit{w/o} flag, where the refinement flag is removed, and the refiner consequently performs refinement at every planning step.

The performance decrease in \textit{actor-only} and \textit{random plan} variants demonstrates the limitation of relying on a single LLM for planning, even when environmental knowledge is available, thus highlighting the advantage of our multi-LLM collaborative paradigm. In addition, the inferior performance of the \textit{w/o} refiner and \textit{w/o} critic variants suggests the necessity of accurate evaluation by the critic and targeted refinement by the refiner to improve the quality of planning. Finally, the performance drop in \textit{w/o} flag reflects the inherent tendency of LLMs towards over-refinement, where satisfactory plans are unnecessarily revised, leading to poor plan quality. Case studies on the over-refinement issue are provided in Appendix \ref{sec:C3}. In contrast, the refinement flag in our method effectively mitigates this issue.

\textbf{Subgoal tracker ablation.} To validate the contribution of the core mechanisms designed in the subgoal tracker, we conduct the following experiments: (1) \textit{extra $r$ every time}, where extra rewards are provided each time a subgoal is achieved; (2) \textit{w/o} extra $r$, where no extra rewards are provided when the agent achieves a subgoal; and (3) \textit{w/o} weight update, where weight update is disabled, thus the subgoal graph remains without weights throughout training.

The results of \textit{extra $r$ every time} show that the agent's behavior deviates from the optimal policy, whereas our design of granting extra rewards upon the first completion of each subgoal in the plan ensures that the RL agent receives additional incentives without compromising optimality. The performance decrease in \textit{w/o} extra $r$ indicates that the RL agent's decisions fail to align with the plan, thereby confirming the effectiveness of extra rewards in explicitly incentivizing the agent to follow the plan. In addition, the results of \textit{w/o} weight update demonstrate that graph weights play a key role in achieving fine-grained improvements in planning quality. Specifically, as illustrated in Figure \ref{fig:hot}, the graph weights explicitly represent the agent's mastery of each subgoal. With weight updates, the LLMs can capture the agent's evolving capabilities during training, enabling the generated plans to better align with execution.

\section{CONCLUSION}
In this work, we introduce SGA-ACR, a novel framework that integrates an environment-specific subgoal graph and entity knowledge base into a multi-LLM planning pipeline, enabling LLM-based planning to generate high-quality plans that effectively guide the RL agent. Furthermore, we propose a subgoal tracker that achieves better alignment between planning and execution in LLM-guided RL. Experimental results in the Crafter environment demonstrate that SGA-ACR not only outperforms baseline methods but also exhibits robustness across different LLM parameter scales, highlighting its potential to address complex open-world environments with any existing LLMs. Despite the advancements of our method, a notable limitation lies in its reliance on detailed environmental information available in advance. Future work could explore learning an accurate subgoal graph from agent-environment interactions.





\bibliographystyle{ACM-Reference-Format} 
\bibliography{sample}


\clearpage
\appendixpage
\appendix
\section{Implementation Details}\label{sec:A}
\subsection{RL Algorithm}

We adopt the PPO algorithm~\cite{schulman2017proximal} for policy learning in SGA-ACR. For the network architecture, a ResNet encoder is employed for image inputs following the IMPALA design~\cite{espeholt2018impala}, while a multilayer perceptron (MLP) is used for text embedding. The detailed implementation follows PPO (ResNet)~\cite{moon2023discovering}. The hyperparameters used for PPO training are summarized in Table \ref{tab:ppo}. Our framework can also be flexibly combined with other RL algorithms.

\begin{table}[ht]
  \centering
  \caption{Hyperparameters for PPO.}
  \label{tab:ppo}
  \begin{tabular}{lc}
    \toprule
    \textbf{Hyperparameter} & \textbf{Value} \\
    \midrule
    Discount factor            & 0.95   \\
    GAE smoothing parameter    & 0.65   \\
    Timesteps per rollout      & 4096   \\
    Epochs per rollout         & 3      \\
    Batch size                 & 512    \\
    Entropy bonus              & 0.01   \\
    PPO clip range             & 0.2    \\
    Learning rate              & 3e-4   \\
    Max grad norm              & 0.5    \\
    Value function coefficient & 0.5    \\
    Extra reward & 0.2    \\
    Optimizer                  & \textit{Adam} \\
    \bottomrule
  \end{tabular}
\end{table}

\subsection{LLM-based Actor-Critic-Refiner Module}
We employ LLMs from the Qwen3 series~\cite{yang2025qwen3technicalreport}. The query parameters for the actor, critic and refiner are summarized in Table \ref{tab:llms}.

\begin{table}[h]
  \centering
  \caption{Hyperparameters for LLMs.}
  \label{tab:llms}
  \begin{tabular}{lc}
    \toprule
    \textbf{Hyperparameter} & \textbf{Value} \\
    \midrule
    Actor temperature   & 0.6 \\
    Critic temperature  & 0.1 \\
    Refiner temperature & 0.2 \\
    Max tokens          & 500 \\
    Query interval      & 100 \\
    Subgoals per plan   &  3  \\
    \bottomrule
  \end{tabular}
\end{table}

\subsection{Full Prompt Details}
In the following, we provide detailed prompts of actor, critic and refiner as well as the verbalized subgoal graph and its description rules.

\noindent\texttt{The prompt of the actor LLM:}
\begin{lstlisting}
# System prompt
You are a professional game analyst for a Minecraft-like game. 
[INPUTS YOU WILL GET]
- Player's state (observation, status, inventory)
- Info of currently observed entities
- The achievements need to be achieved
- The current subgoals available for planning
- The dependency graph between subgoals in a fixed grammar:

[SUBGOAL GRAPH GRAMMAR]
{Graph description}

[YOUR TASK]
- Based on the Subgoal Dependency Graph and the player's state together with other provided information, consider candidate plans that can help the player unlock as many achievements as possible.
- Propose 3 different candidate plans. Each plan should consist of three distinct subgoals (no duplicates inside a plan), and each subgoal must come from ***The Subgoals Available For Planning***.
- For each proposed candidate plan, provide your reason in one clear and concise sentence.

[STRICT RESPONSE FORMAT]:
PlanA<subgoal1,subgoal2,subgoal3>
ReasonA<reason for PlanA>
PlanB<subgoal1,subgoal2,subgoal3>
ReasonB<reason for PlanB>
PlanC<subgoal1,subgoal2,subgoal3>
ReasonC<reason for PlanC>

Make sure to follow the response format strictly! Do not include any extra content beyond what is required!

# User prompt
Player's State: <{text_obs}>
Entity Info: <{entity_info}>
The Achievements Need To Be Achieved: <{unachieved}>
The Subgoals Available For Planning: <{subgoal_set}>
Subgoal Dependency Graph: <{graph_text}>
\end{lstlisting}

\noindent\texttt{The prompt of the critic LLM:}
\begin{lstlisting}
# System prompt
You are a rigorous Minecraft-like game plan critic.

[INPUTS YOU WILL GET]
- Player's state (observation, status, inventory)
- Info of currently observed entities
- The achievements need to be achieved
- The current subgoals available for planning
- The dependency graph between subgoals in a fixed grammar
- Three candidate plans and their reasons provided from the actor
- Detailed information of the subgoals included in the candidate plans

[SUBGOAL GRAPH GRAMMAR]
{Graph description}

[YOUR TASK]
1. Check EACH candidate plan:
- Candidate Validity: Each of 3 subgoals MUST come from ***The Subgoals Available For Planning*** and no duplicates!
- Feasibility and Ordering: Each of 3 subgoals must satisfy the graph-dependency constraints and its prerequisites, match the current capacity of the agent, and follow a correct logical order. Follow the steps below to check prerequisites of each subgoal.
* Step 1: For each subgoal, refer to the Subgoal Dependency Graph and Subgoal Details to identify the preconditions of this subgoal.
* Step 2: Compare the current state with the prerequisites that need to be met, determine whether the prerequisites of this subgoal are all satisfied.
* Step 3: If all prerequisites are satisfied and match the current capacity of the agent, this subgoal is feasible.
- Goal alignment: The plan should focus on unlocking as many achievements as possible.
2. For EACH plan, following the above instructions, provide 4 points of feedback (one clear and concise sentence per point):
- For points 1-3, each point is an evaluation of one subgoal in the plan. In each point:
* First, based on Candidate Validity, determine whether this subgoal is included in The Subgoals Available For Planning (Valid or Invalid).
* If the subgoal is valid, analyze the prerequisites of the subgoal and whether they are satisfied (Feasible or Infeasible). If infeasible, specify which prerequisites are missing. If invalid, end this point and move to the next point.
- The final point should offer the evaluation of Goal alignment for this plan.
3. Based on your feedback, rank the three candidate plans from best to worst, with the best first, using the names of the plans exactly as provided by the actor. e.g., Ranking<PlanB,PlanA,PlanC>
4. Based on your feedback about the top-ranked plan, decide whether the top-ranked plan needs modification:
- Need_Modify<yes>
- Need_Modify<no>

[STRICT RESPONSE FORMAT]
PlanA_feedback<1. ...; 2. ...; 3. ...; 4. ...>
PlanB_feedback<...>
PlanC_feedback<...>
Ranking<name of the best plan,name of the second plan,name of the third plan>
Need_Modify<yes or no for the top-ranked plan>

Make sure to follow the response format strictly! Do not include any extra content beyond what is required!

# User prompt
Player's State: <{text_obs}>
Entity Info: <{entity_info}>
The Achievements Need To Be Achieved: <{unachieved }>
The Subgoals Available For Planning: <{subgoal_text_set}>
Subgoal Dependency Graph: <{graph_text}>
Actor Output: <{actor_output}>
Subgoal Details: <{subgoal_details_text}>
\end{lstlisting}

\noindent\texttt{The prompt of the refiner LLM:}
\begin{lstlisting}
# System prompt
You are an expert refiner for Minecraft-like game plans.

[INPUTS YOU WILL GET]
- Player's state (observation, status, inventory)
- Info of currently observed entities
- The achievements need to be achieved
- The current subgoals available for planning
- The dependency graph between subgoals in a fixed grammar
- The candidate plans
- The feedback of candidate plans from the critic (Ranking is from best to worst and with the best first)

[SUBGOAL GRAPH GRAMMAR]
{Graph description}

[YOUR TASK]
- Based on the provided information, determine whether the top-ranked plan needs to be modified. If no modification is needed, the final plan should remain the same as the top-ranked plan. If modification is required, make the necessary refinement, and the final plan will be the refined plan.
- Based on your analysis, output one final plan. The final plan MUST consist of 3 distinct subgoals and each subgoal must come from ***The Subgoals Available For Planning***. The final plan is to help the player unlock as many achievements as possible.
- First, provide your analysis or reasoning about the final plan in no more than 2 clear and concise points (one sentence per point). Then, output the final plan.

[STRICT RESPONSE FORMAT]
Analysis<1. ...;2. ...>
Final_Plan<subgoal1,subgoal2,subgoal3>
Make sure to follow the response format strictly! Do not include any extra content beyond what is required!

# User prompt
Player's State: <{text_obs}>
Entity Info: <{entity_info}>
The Achievements Need To Be Achieved: <{unachieved}>
The Subgoals Available For Planning: <{subgoal_text_set}>
Subgoal Dependency Graph: <{graph_text}>
Candidate Plans: <{candidate_plans}>
Critic's Feedback: <{critic_feedback}>
\end{lstlisting}

\noindent\texttt{The verbalized subgoal graph:}
\begin{lstlisting}
collect_sapling; collect_water; collect_wood; defeat_skeleton; defeat_zombie; eat_cow; sleep
collect_sapling -> place_plant; collect_wood -> place_table
place_plant -> eat_plant; collect_wood & place_table -> make_wood_pickaxe; collect_wood & place_table -> make_wood_sword
make_wood_pickaxe -> collect_coal; make_wood_pickaxe -> collect_stone; make_wood_sword -> defeat_skeleton; make_wood_sword -> defeat_zombie
collect_stone & collect_wood & place_table -> make_stone_pickaxe; collect_stone & collect_wood & place_table -> make_stone_sword; collect_stone -> place_furnace; collect_stone -> place_stone
make_stone_pickaxe -> collect_iron; make_stone_sword -> defeat_skeleton; make_stone_sword -> defeat_zombie
collect_coal & collect_iron & collect_wood & place_furnace -> make_iron_pickaxe; collect_coal & collect_iron & collect_wood & place_furnace -> make_iron_sword
make_iron_pickaxe -> collect_diamond; make_iron_sword -> defeat_skeleton; make_iron_sword -> defeat_zombie
\end{lstlisting}

\noindent\texttt{The description rules of subgoal graph:}
\begin{lstlisting}
- One line is one depth layer. In each layer, each subgoal on the left of '->' is the prerequisite of the subgoal on the right. Subgoals in higher layer require subgoals in lower layer as prerequisites.
- Item forms:
* ROOT node: the subgoals in the first layer (no prerequisite)
* AND group edge: a & b & c -> x (a & b & c are ALL required for x)
* Single edge: a -> x (any one is sufficient)
- Each root node or edge is followed by a percentage in parentheses, which indicates the current agent's success rate on that root node or edge. (-%) indicates that this root node or edge has not been planned yet.
\end{lstlisting}

\subsection{Text Embedding}
For text embedding, we choose the open-source \textit{all-MiniLM-L6-v2} model as the encoder.

\subsection{Computational resources}
All experiments are conducted on a server equipped with an Intel(R) Xeon(R) Platinum 8280 CPU @ 2.70GHz, an NVIDIA GeForce RTX 3090 GPU, and 256 GB of RAM.

\subsection{Source Code}
For reproducibility, the source code is provided in the supplementary material as \texttt{SGA-ACR.zip}.

\begin{figure*}[ht]
  \centering
  \includegraphics[width=0.8\textwidth]{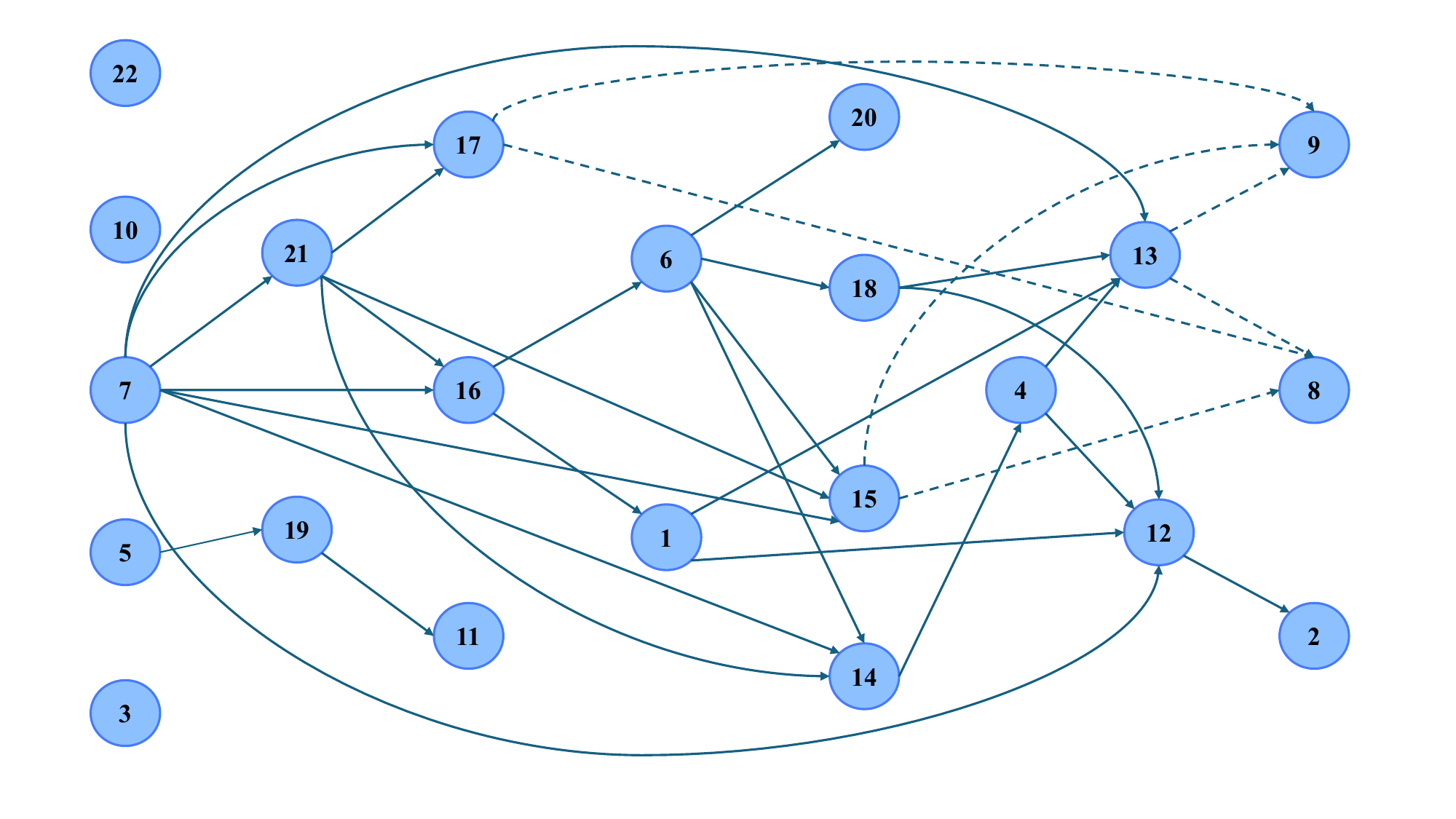}
  \caption{Subgoal graph. Nodes 1-22 correspond to: Collect Coal, Collect Diamond, Collect Drink, Collect Iron, Collect Sapling, Collect Stone, Collect Wood, Defeat Skeleton, Defeat Zombie, Eat Cow, Eat Plant, Make Iron Pickaxe, Make Iron Sword, Make Stone Pickaxe, Make Stone Sword, Make Wood Pickaxe, Make Wood Sword, Place Furnace, Place Plant, Place Stone, Place Table, Sleep. Solid lines denote AND-edges, while dashed lines denote OR-edges.}
  \label{fig:ext-graph}
\end{figure*}

\section{Experiment details}\label{sec:B}
In this section, we present and analyze the results of knowledge extraction and construction in our method (Section \ref{sec:bgkn}), provide the additional learning curves for the main results (Section \ref{sec:mainresult}), and show the corresponding learning curves for the experiments on model-scale analysis (Section \ref{sec:modelscale}) and ablation studies (Section \ref{sec:aba}).

\subsection{Knowledge Extraction and Construction}
The constructed subgoal graph is illustrated in Figure \ref{fig:ext-graph}. As shown, the subgoals in the graph are well aligned with the achievements required in the crafter environment. This demonstrates that the LLM can understand environmental information and task objectives, thereby constructing a comprehensive set of subgoals that satisfies the requirements of high-level planning. Furthermore, by comparing the dependency relations in the subgoal graph with the actual constraints of the environment, we observe that the dependency relations are consistent with the environment dynamics. This indicates that the extracted subgoal graph can accurately model the environment dynamics at the subgoal level. In addition, the entity knowledge base covers 86\% of the entities in the environment. The uncovered entities (\textit{Arrow}, \textit{Path}, and \textit{Lava}) have only minor impact on subgoal-level planning, which confirms the completeness of the knowledge base. In summary, these results validate the accuracy and effectiveness of knowledge extraction and construction in our method.

\subsection{Comparison with Baselines}
\begin{figure*}[ht]
  \centering  
  \includegraphics[width=\textwidth]{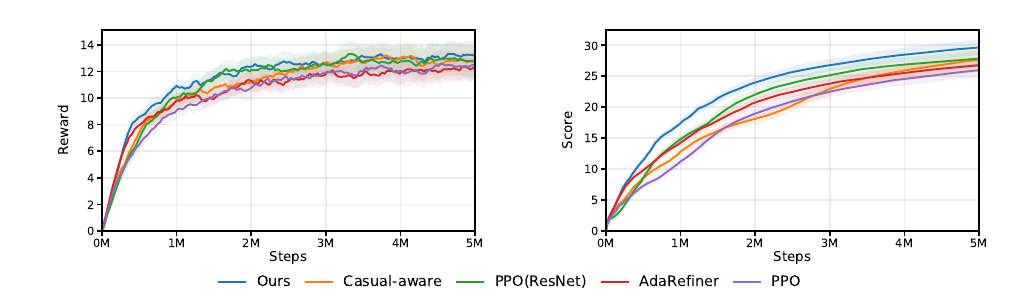}
  \caption{Comparison of reward and score learning curves at 5M steps}
  \label{fig:learning5}
\end{figure*}

In this subsection, we present a comparison of the reward and score learning curves at 5M steps, as shown in Figure \ref{fig:learning5}.

\subsection{Model-Scale Experiments}
\begin{figure*}[ht]
  \centering
  \includegraphics[width=\textwidth]{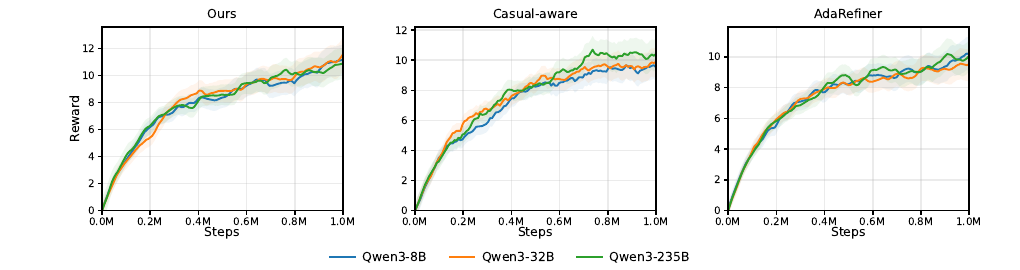}
  \caption{The reward curves of SGA-ACR compared with LLM-guided RL Baselines across various model scales}
  \label{fig:size-rew}
\end{figure*}

\begin{figure*}[ht]
  \centering
  \includegraphics[width=\textwidth]{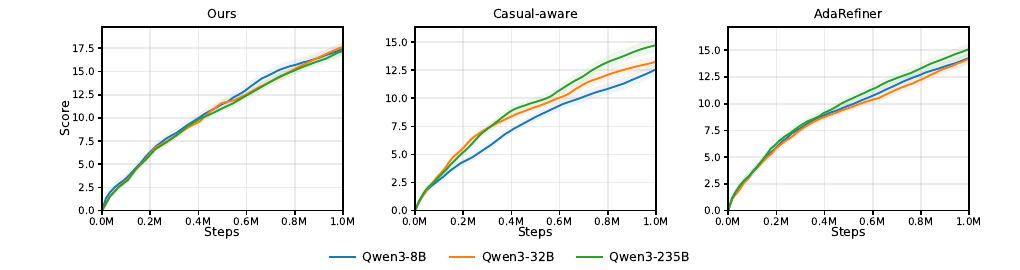}
  \caption{The score curves of SGA-ACR compared with LLM-guided RL Baselines across various model scales}
  \label{fig:size-scr}
\end{figure*}

The experimental results of model scale analysis are presented in Figure \ref{fig:size-rew} and Figure \ref{fig:size-scr}. As shown, our method exhibits consistent learning curves across different model sizes, whereas the LLM-guided RL baselines display substantial variations as the number of model parameters changes. This demonstrates that our method is robust to model scale during training, while the LLM-guided RL baselines are sensitive to parameter size.

\subsection{Ablation Experiments}
\begin{figure*}[ht]
  \centering
  \includegraphics[width=\textwidth]{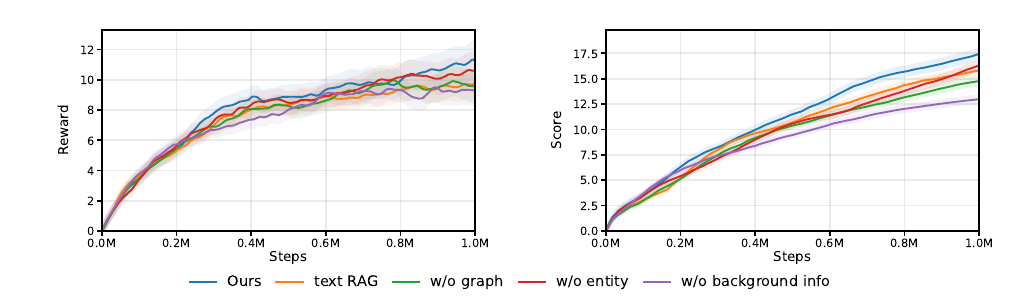}
  \caption{The learning curves of the knowledge module ablation experiments}
  \label{fig:aba-kn}
\end{figure*}

\begin{figure*}[ht]
  \centering
  \includegraphics[width=\textwidth]{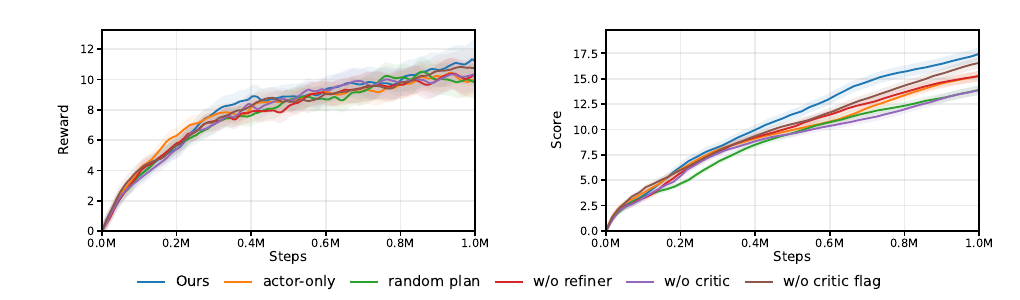}
  \caption{The learning curves of the planning module ablation experiments}
  \label{fig:aba-planning}
\end{figure*}

\begin{figure*}[ht]
  \centering
  \includegraphics[width=\textwidth]{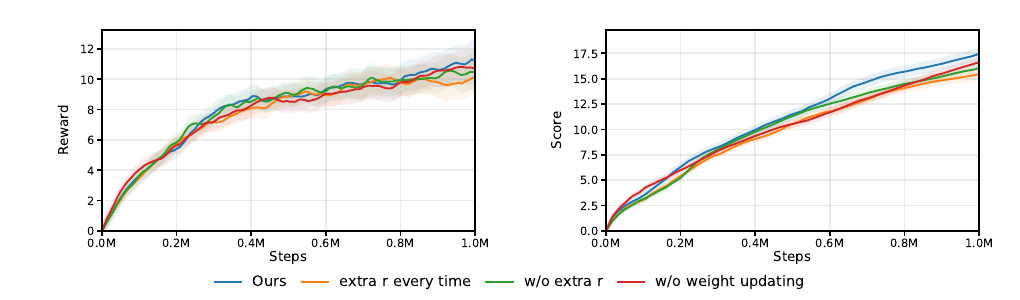}
  \caption{The learning curves of the subgoal tracker ablation experiments}
  \label{fig:aba-tra}
\end{figure*}

The learning curves of the knowledge module ablation experiments are shown in Figure \ref{fig:aba-kn}. The learning curves of the planning module ablation experiments are presented in Figure \ref{fig:aba-planning}, while those of the subgoal tracker ablation experiments are depicted in Figure \ref{fig:aba-tra}. The comparisons of learning curves highlight the positive contribution of each component to the overall performance of our method.

\section{Case Studies}\label{sec:C}
In this section, we analyze the effectiveness of the LLM-based planning module and the ability of the agent to follow subgoals in decision-making through representative test cases. In addition, we examine test cases from the ablation setting \textit{w/o} critic flag to further analyze the over-refinement problem.

\subsection{LLM-based planning Analysis}
\begin{figure*}[ht]
  \centering
  \includegraphics[width=0.8\textwidth]{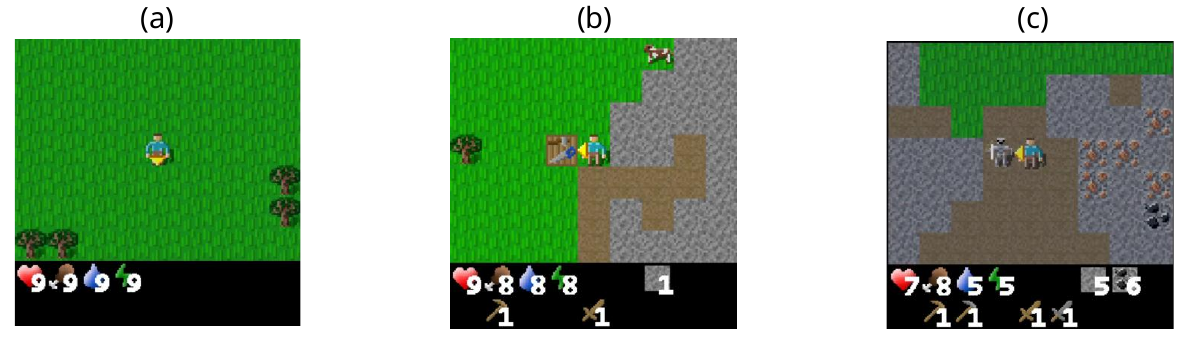}
  \caption{Case details of planning Analysis}
  \label{fig:case_llm}
\end{figure*}

To demonstrate the accuracy of planning in our framework, we analyze the results of three planning steps within a single episode of our method. The game frames at the query time are shown in Figure \ref{fig:case_llm}, and the corresponding outputs of the planning LLMs are provided as follows.

The outputs of the planning module indicate that the actor can generate multiple feasible planning paths based on the dependency relations in the subgoal graph. The critic, by leveraging detailed information of the nodes along each path, provides accurate and fine-grained evaluations of the candidate plans. When necessary, the refiner integrates and refines these evaluations to produce the final plan. This planning process, grounded in structured environmental knowledge, significantly enhances the effectiveness and reliability of the generated plans.

\noindent\texttt{The output of the planning module corresponding to Figure \ref{fig:case_llm}(a):}
\begin{lstlisting}
# actor output
PlanA<collect_wood,place_table,make_wood_sword>
ReasonA<Collecting wood and placing a table are prerequisites for making a wood sword, which can help defeat zombies and skeletons, unlocking multiple achievements.> 
PlanB<collect_stone,place_furnace,make_stone_sword>
ReasonB<Collecting stone and placing a furnace are prerequisites for making a stone sword, which can help defeat zombies and skeletons, unlocking multiple achievements.>
PlanC<collect_sapling,place_plant,eat_plant>
ReasonC<Collecting saplings and placing plants are prerequisites for eating plants, which directly unlocks the eat_plant achievement.>"

# critic output
PlanA_feedback<1. collect_wood is valid and feasible as the player sees a tree and can gather wood without tools. 2. place_table is valid and feasible as the player can gather wood (from collect_wood) and place a table without tools. 3. make_wood_sword is valid and feasible as the prerequisites (collect_wood and place_table) are met. 4. This plan aligns well with the goal of making a wood sword to defeat zombies and skeletons, unlocking multiple achievements.>
PlanB_feedback<1. collect_stone is invalid as the player does not have a wood_pickaxe, which is required to mine stone. 2. place_furnace is valid and feasible as the player can collect stone (once wood_pickaxe is made) and place a furnace. 3. make_stone_sword is valid and feasible as the prerequisites (collect_stone, collect_wood, and place_table) are met. 4. This plan aligns well with the goal of making a stone sword to defeat zombies and skeletons, unlocking multiple achievements.>
PlanC_feedback<1. collect_sapling is valid and feasible as the player sees grass and can collect saplings without tools. 2. place_plant is valid and feasible as the player can plant a sapling (from collect_sapling) and place a plant. 3. eat_plant is valid and feasible as the player can eat a ripe plant (from place_plant). 4. This plan aligns well with the goal of eating plants, directly unlocking the eat_plant achievement.> 
Ranking<PlanA,PlanC,PlanB>
Need_Modify<no>
\end{lstlisting}

\noindent\texttt{The output of the planning module corresponding to Figure \ref{fig:case_llm}(b):}
\begin{lstlisting}
# actor output 
PlanA< collect_wood make_wood_sword,collect_wood,defeat_zombie >
ReasonA<Defeating zombie and skeleton can be achieved by making a wood sword, which requires collecting wood, aligning with the dependency graph.>
PlanB<collect_stone,make_stone_pickaxe,make_stone_sword>
ReasonB<Collecting stone is a prerequisite for making both stone pickaxe and stone sword, which are key to defeating enemies and progressing.>
PlanC<eat_cow,make_wood_sword,defeat_zombie>
ReasonC<Eating cow provides food, and making a wood sword enables defeating zombie, which are both critical for survival and progression.>

# critic output
PlanA_feedback<1. collect_wood is valid and feasible as the player sees a tree and can collect wood without tools. 2. make_wood_sword is valid and feasible as the player has a wood_sword in inventory and can craft it at the table. 3. defeat_zombie is valid but infeasible as the player does not have a sword (wood_sword is already in inventory, so this is feasible). 4. This plan aligns well with the goal of defeating zombies and using the wood sword, but it does not contribute to other critical achievements like collecting resources or crafting tools.>
PlanB_feedback<1. collect_stone is valid but infeasible as the player does not have a wood_pickaxe (already in inventory, so this is feasible). 2. make_stone_pickaxe is valid but infeasible as the player needs wood and stone (already has wood and stone, but needs to place table). 3. make_stone_sword is valid but infeasible as the player needs wood and stone (already has both, but needs to place table). 4. This plan aligns well with the goal of making stone tools and defeating enemies, but it requires placing a table which is not yet done.>
PlanC_feedback<1. eat_cow is valid and feasible as the player sees a cow and can eat it. 2. make_wood_sword is valid and feasible as the player has a wood_sword in inventory and can craft it at the table. 3. defeat_zombie is valid but infeasible as the player does not have a sword (but wood_sword is already in inventory). 4. This plan aligns well with the goal of eating cow and defeating zombies, but it does not contribute to other critical achievements like collecting resources or crafting tools.>
Ranking<PlanB,PlanC,PlanA>
Need_Modify<no>
\end{lstlisting}

\noindent\texttt{The output of the planning module corresponding to Figure \ref{fig:case_llm}(c):}
\begin{lstlisting}
# actor output
PlanA<collect_iron,make_iron_pickaxe,make_iron_sword>
ReasonA<This plan focuses on quickly moving toward iron tool crafting, which is powerful in the long run but may not be feasible due to missing prerequisites like furnace placement and wood.>
PlanB<collect_sapling,place_plant,eat_plant>
ReasonB<This plan emphasizes survival through food gathering, useful for sustaining energy, but does not contribute directly to tool progression.>
PlanC<make_stone_sword,defeat_zombie,collect_stone>
ReasonC<This plan emphasizes combat with stone weapons and gathering additional stone, which supports survival and building but does not advance toward unlocking iron collection.>

# critic output
PlanA_feedback<1. collect_iron: Valid. Prerequisites: stone_pickaxe (already in inventory), iron (visible). Feasible. 2. make_iron_pickaxe: Valid. Prerequisites: wood (missing), coal (6), iron (not yet collected), place_furnace (not placed). Infeasible. 3. make_iron_sword: Valid. Prerequisites: wood (missing), coal (6), iron (not yet collected), place_furnace (not placed). Infeasible. 4. PlanA aligns with long-term achievements but lacks foundational prerequisites like furnace and wood.>
PlanB_feedback<1. collect_sapling: Valid. Prerequisites: grass (visible). Feasible. 2. place_plant: Valid. Prerequisites: sapling (not in inventory). Infeasible. 3. eat_plant: Valid. Prerequisites: grown plant (not present). Infeasible. 4. PlanB aligns with food-related achievements but contributes little to combat or resource progression.>
PlanC_feedback<1. make_stone_sword: Valid. Prerequisites: wood (missing), stone (available), place_table (not placed). Infeasible. 2. defeat_zombie: Valid. Prerequisites: at least a wood/stone sword (already have both). Feasible. 3. collect_stone: Valid. Prerequisites: wood_pickaxe or stone_pickaxe (already in inventory). Feasible. 4. PlanC aligns with combat readiness and resource gathering but misses the furnace/iron path to unlock deeper progression.>
Ranking<PlanA,PlanC,PlanB>
Need_Modify<yes>

# refiner output 
Analysis<1. PlanA includes iron but skips necessary prerequisites like furnace. 2. PlanB focuses on food and survival, not progression, and PlanC covers combat with zombies but does not address furnace/iron dependencies.>
Final_Plan<defeat_skeleton,place_furnace,collect_iron>
\end{lstlisting}

\subsection{Agent Behavior Analysis}
\begin{figure*}[ht]
  \centering
  \includegraphics[width=0.8\textwidth]{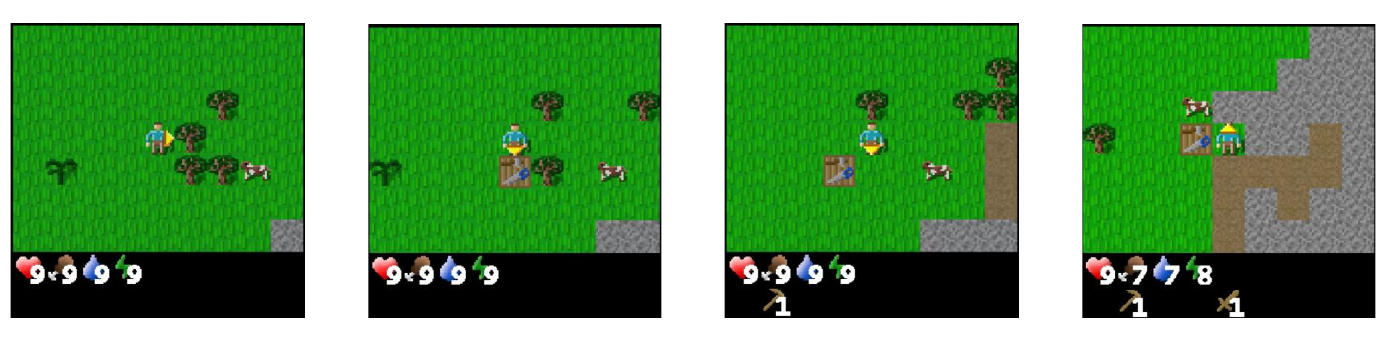}
  \caption{Case details of making wood tools.}
  \label{fig:case_wood}
\end{figure*}

\begin{figure*}[ht]
  \centering
  \includegraphics[width=0.8\textwidth]{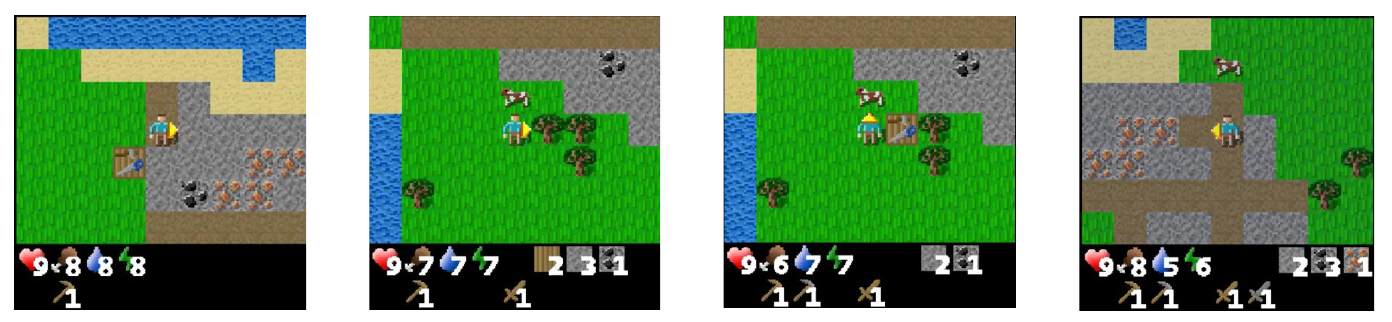}
  \caption{Case details of making stone tools.}
  \label{fig:case_stone}
\end{figure*}

To investigate the agent’s ability to follow subgoals in decision-making, we analyze key frames from episodes where the subgoals involve making wooden and stone tools. As shown in Figure \ref{fig:case_wood}, for the subgoals \textit{Make Wood Pickaxe} and \textit{Make Wood Sword}, the agent first collects two pieces of wood to craft a workbench, then gathers one additional piece of wood to craft a wooden pickaxe near the workbench, and subsequently collects another piece of wood to complete the wooden sword. Similarly, as illustrated in Figure \ref{fig:case_stone}, for the subgoals \textit{Make Stone Pickaxe} and \textit{Make Stone Sword}, once the agent has obtained a wood pickaxe, it begins to collect stone while simultaneously gathering wood. Upon accumulating two pieces of wood, the agent immediately crafts a workbench, and then continues to gather the additional wood required for stone tools. After meeting the resource requirements, the agent returns to the workbench to complete the crafting of both the stone pickaxe and stone sword. These behaviors demonstrate that the agent not only follows high-level subgoals in its decision-making but also accurately satisfies the prerequisite conditions for each subgoal before achieving them. This further validates the advantage of combining LLM-based high-level guidance with RL, as well as the effectiveness of our subgoal tracker in aligning planning with execution.

\subsection{Over-Refinement Analysis} \label{sec:C3}
\begin{figure*}[ht]
  \centering
  \includegraphics[width=0.6\linewidth]{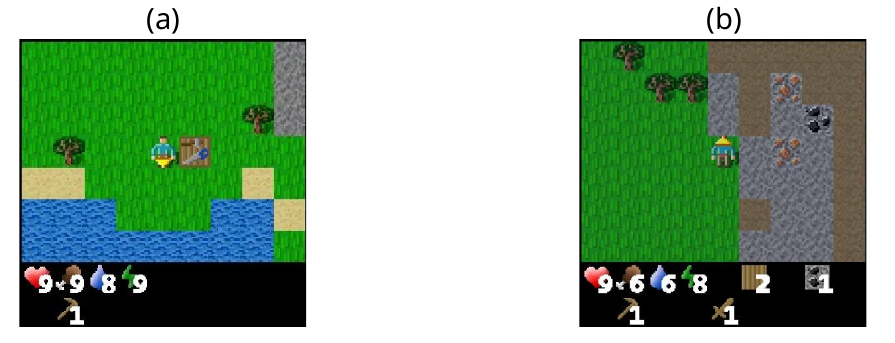}
  \caption{Case details of over-refinement.}
  \label{fig:case_refine}
\end{figure*}

To illustrate the over-refinement problem that arises when the Actor-Critic-Refiner framework does not explicitly determine whether refinement is necessary, we analyze two planning cases from the \textit{w/o} critic flag experiments. The game frames at the planning time are illustrated in Figure \ref{fig:case_refine} and the corresponding outputs of the planning module are shown as follows.

\noindent\texttt{The output of the planning module corresponding to Figure \ref{fig:case_refine}(a):}
\begin{lstlisting}
# actor output
PlanA<collect_wood,make_wood_sword,collect_stone>
ReasonA<Tree is visible so wood is obtainable; with wood and a table in view a wood sword can be crafted for safety; the existing wood_pickaxe then enables collecting stone efficiently.>
PlanB<collect_stone,place_furnace,make_stone_pickaxe>
ReasonB<Push tech progression by gathering stone first, setting a furnace for future smelting, and then crafting a stone pickaxe.>
PlanC<collect_wood,place_table,make_wood_pickaxe>
ReasonC<Reinforce the basic toolchain by ensuring table access and maintaining wood-pickaxe capability, though it advances fewer new achievements.>

# critic output
PlanA_feedback<1. collect_wood: Valid and feasible because a tree is in view and wood needs no tools; 2. make_wood_sword: Valid and feasible since prerequisites (collect_wood and access to table) are satisfied with a table visible; 3. collect_stone: Valid and feasible given the player already has a wood_pickaxe; 4. This plan improves combat safety and immediately unlocks the stone branch with minimal steps.>
PlanB_feedback<1. collect_stone: Valid and feasible via the current wood_pickaxe; 2. place_furnace: Valid but may be infeasible immediately if insufficient stone is gathered after step 1; 3. make_stone_pickaxe: Valid yet longer chain because it still requires wood and table access; 4. This plan prioritizes tech but delays safety and adds extra steps.>
PlanC_feedback<1. collect_wood: Valid and feasible; 2. place_table: Valid and feasible but redundant given a table is already in view; 3. make_wood_pickaxe: Valid yet redundant because a wood_pickaxe is already in inventory; 4. This plan offers little progress toward unachieved milestones compared with PlanA.>
Ranking<PlanA,PlanB,PlanC>

# refiner output
Analysis<1. PlanA excels by combining early safety (wood sword) with quick access to the stone branch; 2. PlanB highlights longer-term tech via an early furnace, attempting to merge both strengths.>
Final_Plan<collect_wood,collect_stone,place_furnace >
\end{lstlisting}

\noindent\texttt{The output of the planning module corresponding to Figure \ref{fig:case_refine}(b):}
\begin{lstlisting}
# actor output
PlanA<place_table,collect_stone,make_stone_pickaxe>
ReasonA<Use current wood to place a table, mine stone with the existing wood_pickaxe, then craft a stone pickaxe (requires stone + wood + table).>
PlanB<collect_stone,place_furnace,make_stone_pickaxe>
ReasonB<Push tech progression by gathering stone, setting up a furnace for future smelting, and then crafting a stone pickaxe.>
PlanC<defeat_skeleton,defeat_zombie,make_wood_sword>
ReasonC<Attempt nearby combat achievements while maintaining a basic weapon for safety, though it adds little toward tool upgrades.>

# critic output
PlanA_feedback<1. place_table: Valid and feasible since the player has wood in inventory and the graph allows placing a table from collect_wood; 2. collect_stone: Valid and feasible because the player already has a wood_pickaxe and stone is visible; 3. make_stone_pickaxe: Valid and feasible after steps 1 and 2, as prerequisites (stone, wood, table) are then satisfied; 4. This plan uses minimal steps to unlock the stone-tool branch and enable iron collection next.>
PlanB_feedback<1. collect_stone: Valid and feasible with the current wood_pickaxe; 2. place_furnace: Valid but may be infeasible immediately if insufficient stone is gathered after step 1; 3. make_stone_pickaxe: Valid but infeasible in this order because it also requires a placed table, which this plan omits; 4. This plan delays the core upgrade and risks resource shortfall.>
PlanC_feedback<1. defeat_skeleton: Valid and potentially feasible with the existing wood_sword, but risky given mid-level food/drink; 2. defeat_zombie: Same concerns as 1, offering little tech progression; 3. make_wood_sword: Valid yet redundant because a wood_sword is already in inventory; 4. This plan focuses on combat without advancing toward stone/iron milestones.>
Ranking<PlanA,PlanB,PlanC>

# refiner output
Analysis<1. PlanA excels by meeting all prerequisites in minimal steps-place the table, mine stone, then craft a stone pickaxe; 2. PlanB highlights longer-term tech progression via an early furnace>
Final_Plan<collect_stone,place_furnace,make_stone_pickaxe>
\end{lstlisting}

These cases reveal that even when the critic gives highly favorable evaluations to the top-ranked plan, the refiner still tends to modify it. Such modifications often incorrectly integrate subgoals from other candidate plans, which degrades an already optimal plan. In contrast, by introducing the critic flag in our method, the critic is explicitly prompted to reason further based on its own feedback, thereby clarifying the necessity of refinement and significantly alleviating the over-refinement problem.

\end{document}